\documentclass[pmlr]{jmlr}

\usepackage{adjustbox}

\usepackage{longtable}

\usepackage{booktabs}
\usepackage[load-configurations=version-1]{siunitx} 
 \usepackage{cleveref}

\usepackage{tikz}

\theorembodyfont{\upshape}
\theoremheaderfont{\scshape}
\theorempostheader{:}
\theoremsep{\newline}

\newcommand{\pr}[1]{\left(#1 \right)} 

\jmlrvolume{}
\firstpageno{1}
\editors{List of editors' names}

\jmlryear{2024}
\jmlrworkshop{Symmetry and Geometry in Neural Representations}

\title[]{Symmetry-Aware Generative Modeling through Learned Canonicalization}

  \author{\Name{Kusha Sareen} \Email{kusha.sareen@mila.quebec}\\
  \Name{Daniel Levy} \Email{daniel.levy@mila.quebec}\\
  \Name{Arnab Kumar Mondal} \Email{arnab.mondal@mila.quebec}\\
  \Name{Sékou-Oumar Kaba} \Email{kabaseko@mila.quebec}\\
  \Name{Tara Akhound-Sadegh} \Email{tara.akhoundsadegh@mila.quebec}\\
  \Name{Siamak Ravanbakhsh} \Email{siamak.ravanbakhsh@mila.quebec}\\
  \addr Mila, McGill University}

\begin{document}

\maketitle

\begin{abstract}
Generative modeling of symmetric densities has a range of applications in AI for science, from drug discovery to physics simulations.  The existing generative modeling paradigm for invariant densities combines an invariant prior with an equivariant generative process. 
However, we observe that this technique is not necessary and has several drawbacks resulting from the limitations of equivariant networks.
Instead, we propose to model a learned slice of the density so that only one representative element per orbit is learned.
To accomplish this, we learn a group-equivariant canonicalization network that maps training samples to a canonical pose and train a non-equivariant generative model over these canonicalized samples.
We implement this idea in the context of diffusion models. Our preliminary experimental results on molecular modeling are promising,  demonstrating improved sample quality and faster inference time.
\end{abstract}

\section{Introduction}
\label{sec:intro}


Equivariant models have emerged as a popular family of models for generative modeling tasks with symmetry, see e.g. \citep{kohler2020equivariant, shi2021learning, edm, midi,bose2024sestochastic}. These models are especially prevalent in physical domains, which are subject to Euclidean symmetries (rotations and translations) and permutation symmetries.

However, equivariant models have some limitations. One issue is computational complexity, as specialized operations like group convolutions or spherical harmonics are expensive~\citep{kondor2018clebsch, cohen2018spherical, weiler2019general}. 
Additionally, there are expressivity constraints, since limiting models to equivariant functions may restrict their ability to capture complex patterns not perfectly aligned with the symmetry group~\citep{maron2019universality, ravanbakhsh2017equivariance, zhou2020group}. There are generally trade-offs between expressivity and computational cost with equivariant models \citep{maron19a,ravanbakhsh20a, joshi23a, xie2024the}. Furthermore, they can be cumbersome to design and utilize. More flexible models with fewer rigid assumptions are expected to scale better as available compute grows.

Despite these shortcomings, we see two common motivations for imposing equivariance in generative models.
First, using an equivariant generator guarantees that we model an invariant distribution if we start from an invariant prior distribution \citep{kohler2020equivariant, xu2022geodiffgeometricdiffusionmodel}. Second, equivariance can serve as a useful inductive bias for the generator: if a data distribution over $X$ is invariant to a group $G$, then an equivariant model can generalize what it learns for sample $x \in X$ to $g \cdot x, \forall g\in G$  making it more sample efficient.

However, for many tasks, having an invariant output distribution isn't necessary. Consider molecules: generating novel and stable molecules is more important than ensuring molecules can be generated in all orientations.
 Instead of modelling the density on $X$, it suffices to model over $X/G$, the set of orbits of $X$ with respect to a group $G$. To do this we need an orbit representative map $X/G \rightarrow X$ that maps all samples in a given orbit to a single representative. Then, we can model a distribution on $X/G$ using any non-equivariant generator, giving the added benefit of increased flexibility and speed.
Following recent works \citep{kaba2023equivariance, kim2023, mondal2024equivariant, allingham2024generative}, we accomplish this by using a canonicalization function, which transforms data into a standardized or ``canonical" form, which is a representative element in its orbit under the group.
By parameterizing $c$ as a neural network, the model can discover an optimal way to standardize data, leading to improved representations and performance. This maintains the inductive bias benefit of equivariance by ensuring similarly structured data is oriented similarly for the generator.

We demonstrate this approach in our experiments using denoising diffusion models \citep{ho2020denoising}.
We show on a molecular dataset that using a simple non-equivariant denoising network along with a learned canonicalizer results in higher-quality samples than using an equivariant network, while halving inference time.

\begin{figure}
    \centering
    \includegraphics[width=0.9\linewidth]{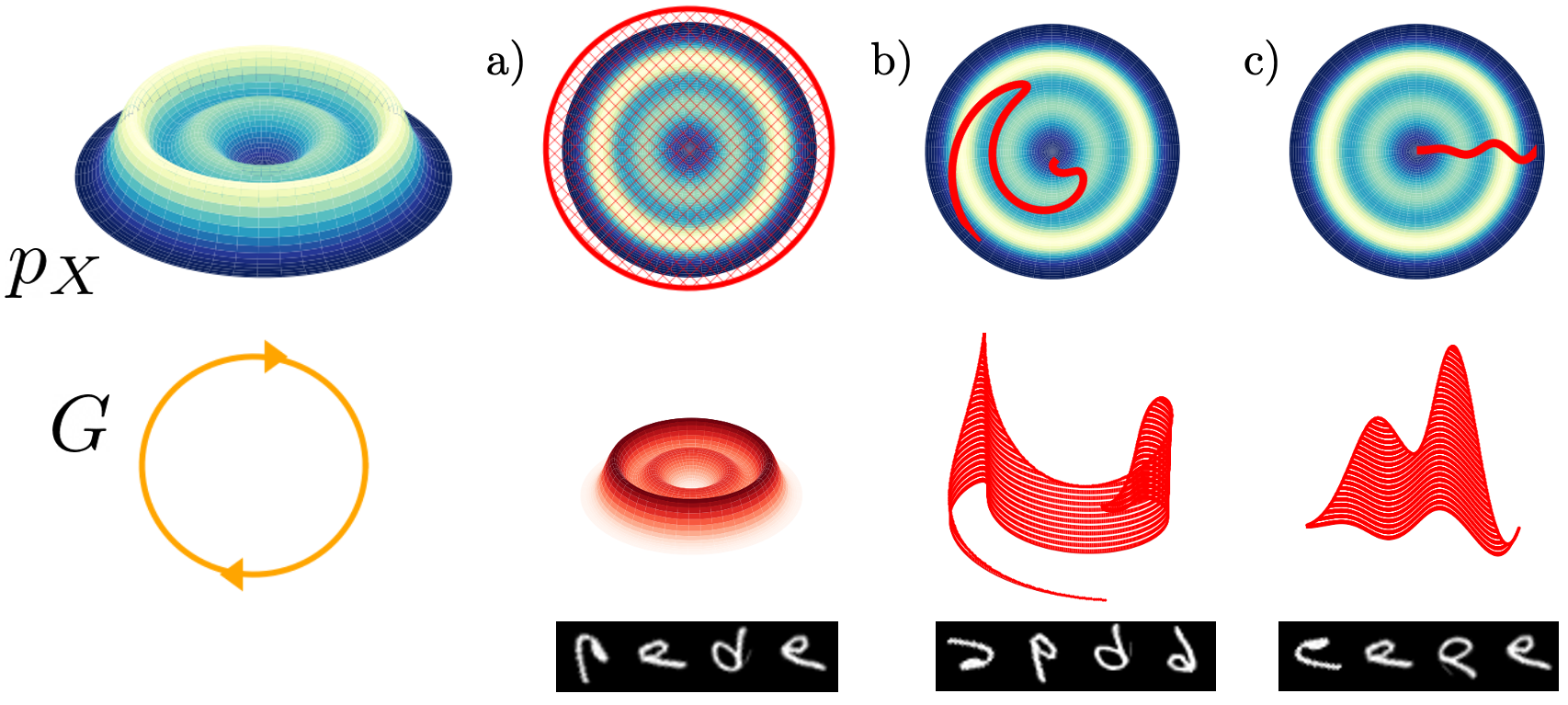}
    \caption{\footnotesize{Left: A distribution $p_X$ that's invariant to the group of rotations $G$.
    Right: a) A non-equivariant generative model must learn the whole distribution $p_X$. b) A fixed canonicalization method outputs group elements that map different samples of $p_X$ onto a single slice (red), which may be learned by a generative model. c) A learned canonicalizer can map samples onto a slice that yields a simpler distribution for the generative model to learn.}}
    \label{fig:main}
    \vspace{-1em}
\end{figure}

\vspace{-2ex}

\section{Methods}
\label{sec:methods}

Our goal is to learn a distribution over the orbits $X/G$ instead of directly learning a distribution over $X$. For this, we define the \textit{orbit representative map} $c: X \rightarrow X/G$ as $c\pr{x} =  h\pr{x}^{-1} x$, following \cite{bloem2020probabilistic}. The function $h: X \rightarrow G$ is a relaxed equivariant canonicalization function \cite{kaba2023symmetry}. The invariant orbit representative map chooses a representative from the orbit $G\cdot x$ of any sample $x$.

Given the map $c$ and a data distribution $p_X(x)$, we then consider modelling the distribution over the orbit representatives as $p_{X/G}\pr{c\pr{x}}$.
This essentially corresponds to projecting the distributions over slices defined by orbit-representatives, as shown in \figureref{fig:main}. The distribution $p_{X/G}$ is then modelled using a non-equivariant generator network $\phi$. The resulting training procedure for a denoising diffusion model is described in \algorithmref{alg:train}.
\begin{algorithm2e}
\caption{Training a denoiser to produce canonical samples.}\label{alg:train}
\textbf{Input}: Data point $x$, denoising network $\phi$, equivariant network $h$ \;
Define orbit representative map $c: X/G \rightarrow X$ by $c(x) = h(x)^{-1} x$\;
While not converged:\\
\hspace{10pt}Sample $t \sim U(0, ..., T), \epsilon \sim \mathcal{N}(0, I)$\;
\hspace{10pt}Compute $ z_t = \alpha_t c(x) + \sigma_t \epsilon $ \;
\hspace{10pt}Update $h$, $\phi$ to minimize $\| \epsilon - \phi(z_t, t)\|^{2}$\;
\end{algorithm2e}

The central hypothesis of our framework is that projecting the distribution over orbits should result in a considerably simpler learning problem compared to modelling the full distribution. Furthermore, learning the quotient map via the canonicalization function $h$, should slice the distribution in a way that is easier to model for the generator $\phi$.

It is common practice to set the center of mass to be zero when modelling 3D molecules in order to be invariant to translations, as one cannot define a translationally-invariant prior distribution in Euclidean space \citep{kohler2020equivariant}.  Our method can be seen as a generalization of the center of this mass removal procedure to any group, and we therefore similarly circumvent the issue of needing an invariant prior.





\section{Experiments}
\label{sec:results}

\begin{figure}
    \centering
    \includegraphics[width=0.9\linewidth]{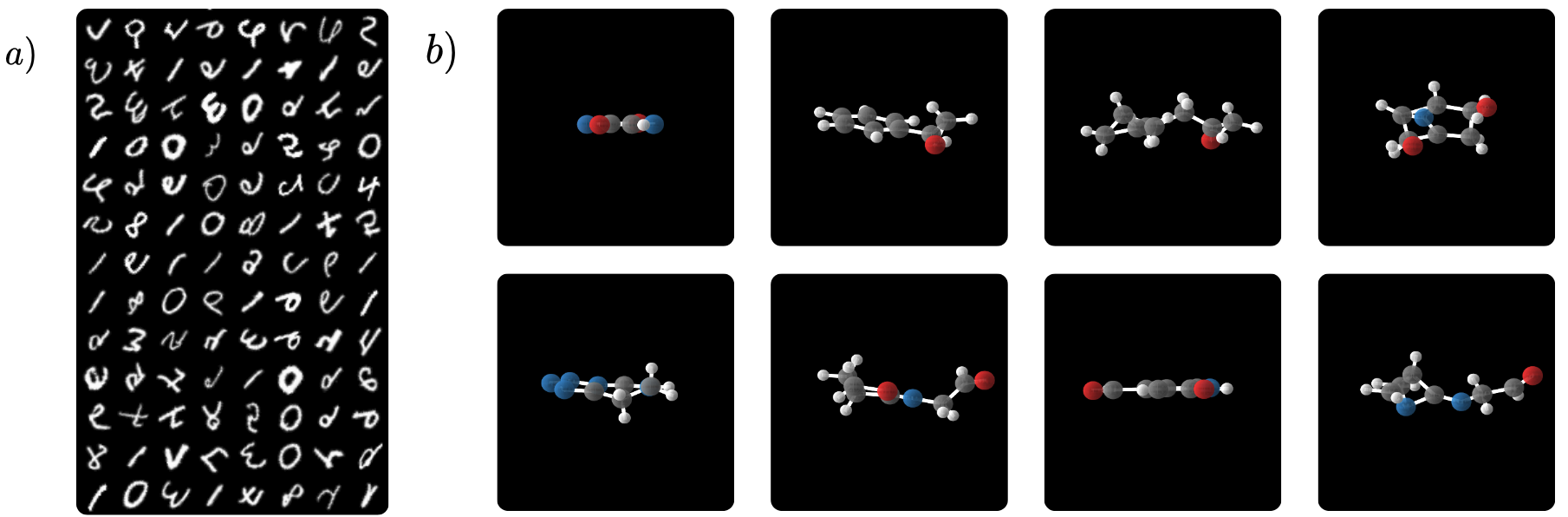}
    \caption{\footnotesize{a) Output of the learned canonicalizer on RotMNIST after training showing aligned digits. b) Select molecules generated from canon+GDM in a near-canonical pose.}}
    \label{fig:samples}
\end{figure}

\subsection{Rotated MNIST Generation}
We start with a simple task on images to demonstrate our framework. A UNet \citep{ronneberger2015unetconvolutionalnetworksbiomedical} is trained to denoise rotated MNIST digits \citep{rotmnsit}, a common classification benchmark for equivariant architechtures. We then train the same network with discrete image canonicalization using an ESCNN \citep{e2cnn} with 8, 16 and 32 discretizations. Results in \tableref{tab:rotmnist} (\cref{apd:first}) show an improved loss on the test set after canonicalization is applied. This method gives some visual intuition for what canonicalization is doing. \figureref{fig:samples} shows the output of a trained canonicalization network. Digits across all classes are aligned, resulting in an easier task for the denoiser.

\subsection{QM9 Molecule Generation}
\textbf{Experiment Setup.}
Models are trained to generate molecules in the QM9 dataset \citep{qm9}. As baselines, an Equivariant Diffusion Model (EDM) \citep{edm} and non-equivariant GNN Diffusion Model (GDM) with identical hyperparameters are trained on the task. Then, a multi-channel EGNN \citep{levy2023using, satorras2021n} is introduced as the canonicalization network, outputting a rotation matrix that is applied to the molecule before denoising with the GDM backbone (canon+GDM). We show that even an unlearned canonicalizer with frozen weights (canon(fr)+GDM) aids the GDM in producing stable molecules.

\textbf{Evaluation Setup.}
A similar evaluation setup to \citet{satorras2022enequivariantnormalizingflows} and \citet{edm} is used. We report the test NLL in addition to a number of desirable features of generated molecules relevant to the drug discovery pipeline: atom stability, molecule stability, validity, uniqueness and inference time.

\textbf{Results.} Our findings (\tableref{qm9results}) show that canon+GDM outperforms EDM across several metrics while halving inference time. Additionally, the unlearned canonicalizer, canon(fr)+GDM, is still able to generate stable molecules with results similar to EDM. \figureref{fig:samples} showsf samples from the canon+GDM model. These molecules are generated in a near-canonical orientation: planar molecules are entirely confined to the xy plane while other oblong molecules are primarily horizontal rather than vertical. This supports the notion that the canonicalizer is able to learn to transform to poses that make denoising easier and that the denoiser is able to generate molecules in such poses.

Although we used a simple non-equivariant counterpart of the equivariant generative model from \cite{hoogeboom2022equivariant} as a denoising network in this experiment, our flexible method can accommodate any arbitrarily expressive non-equivariant model.


\begin{table}
  \caption{\footnotesize{QM9 Results.}}
  \vspace{1em}
  \label{qm9results}
  \centering
  \begin{adjustbox}{width=0.95\textwidth}
  \begin{tabular}{lllllll}
    \toprule
    \cmidrule(r){1-3}
    Model     & NLL & Mol stable   & At stable & Valid & Unique & s/sample\\
    \midrule
    GDM     & $-105.8$ & 70.6     & 97.5 & 88.6 & 99.8 & \textbf{0.55}\\
    EDM     & $-110.7 \pm 1.7$ & 82.0 $\pm$ 0.4     & \textbf{98.7} $\pm$ 0.1 & 91.9 $\pm$ 0.5 & 90.7 $\pm$ 0.6 & 1.16\\
    canon(fr)+GDM     & $-104.1$ & 82.1    & 97.8 & 92.9 & 99.5 & \textbf{0.55}\\
    canon+GDM     & \textbf{-117.4} $ \pm 1.2$ & \textbf{84.6} $\pm$ 0.4  & 98.4 $\pm$ 0.2 & \textbf{94.4} $\pm$ 0.3 & 99.7 $\pm$ 0.2 & \textbf{0.55}\\
    \bottomrule
  \end{tabular}
  \end{adjustbox}
\end{table}

\section{Conclusion}
\label{sec:conclusion}
We show that contrary to previous work, modelling the entire invariant distribution $p_X$ is unnecessary and simply modeling $p_{X/G}$ via canonicalization is a more tractable task. Using a learned canonicalizer allows us to maintain the inductive bias benefit of equivariant neural networks while gaining flexibility and efficiency by the use of a fast expressive non-equivariant generator. In future work, we expect the flexibility of this method will allow for the use of even more powerful generators for generative modelling tasks across domains with symmetry.
    
\bibliography{pmlr-sample}

@misc{edm,
      title={Equivariant Diffusion for Molecule Generation in 3D}, 
      author={Emiel Hoogeboom and Victor Garcia Satorras and Clément Vignac and Max Welling},
      year={2022},
      eprint={2203.17003},
      archivePrefix={arXiv},
      primaryClass={cs.LG},
      url={https://arxiv.org/abs/2203.17003}, 
}

@InProceedings{joshi23a,
  title = 	 {On the Expressive Power of Geometric Graph Neural Networks},
  author =       {Joshi, Chaitanya K. and Bodnar, Cristian and Mathis, Simon V and Cohen, Taco and Lio, Pietro},
  booktitle = 	 {Proceedings of the 40th International Conference on Machine Learning},
  pages = 	 {15330--15355},
  year = 	 {2023},
  editor = 	 {Krause, Andreas and Brunskill, Emma and Cho, Kyunghyun and Engelhardt, Barbara and Sabato, Sivan and Scarlett, Jonathan},
  volume = 	 {202},
  series = 	 {Proceedings of Machine Learning Research},
  month = 	 {23--29 Jul},
  publisher =    {PMLR},
  url = 	 {https://proceedings.mlr.press/v202/joshi23a.html}
}

@misc{midi,
      title={MiDi: Mixed Graph and 3D Denoising Diffusion for Molecule Generation}, 
      author={Clement Vignac and Nagham Osman and Laura Toni and Pascal Frossard},
      year={2023},
      eprint={2302.09048},
      archivePrefix={arXiv},
      primaryClass={cs.LG},
      url={https://arxiv.org/abs/2302.09048}, 
}

@misc{equiadapt,
      title={Equivariant Adaptation of Large Pretrained Models}, 
      author={Arnab Kumar Mondal and Siba Smarak Panigrahi and Sékou-Oumar Kaba and Sai Rajeswar and Siamak Ravanbakhsh},
      year={2023},
      eprint={2310.01647},
      archivePrefix={arXiv},
      primaryClass={cs.LG},
      url={https://arxiv.org/abs/2310.01647}, 
}

@article{kaba2023symmetry,
	author = {Kaba, S{\'e}kou-Oumar and Ravanbakhsh, Siamak},
	journal = {arXiv preprint arXiv:2312.09016},
	title = {Symmetry Breaking and Equivariant Neural Networks},
	year = {2023}}

@article{allingham2024generative,
  title={A Generative Model of Symmetry Transformations},
  author={Allingham, James Urquhart and Mlodozeniec, Bruno Kacper and Padhy, Shreyas and Antor{\'a}n, Javier and Krueger, David and Turner, Richard E and Nalisnick, Eric and Hern{\'a}ndez-Lobato, Jos{\'e} Miguel},
  journal={arXiv preprint arXiv:2403.01946},
  year={2024}
}

@inproceedings{sohl2015deep,
  title={Deep unsupervised learning using nonequilibrium thermodynamics},
  author={Sohl-Dickstein, Jascha and Weiss, Eric and Maheswaranathan, Niru and Ganguli, Surya},
  booktitle={International Conference on Machine Learning},
  pages={2256--2265},
  year={2015}
}

@inproceedings{ho2020denoising,
  title={Denoising diffusion probabilistic models},
  author={Ho, Jonathan and Jain, Ajay and Abbeel, Pieter},
  booktitle={Advances in Neural Information Processing Systems},
  pages={6840--6851},
  year={2020}
}

@inproceedings{kondor2018clebsch,
  title={Clebsch--gordan nets: a fully fourier space spherical convolutional neural network},
  author={Kondor, Risi and Lin, Zhen and Trivedi, Shubhendu},
  booktitle={Advances in Neural Information Processing Systems},
  volume={31},
  pages={10117--10126},
  year={2018}
}

@article{thomas2018tensor,
  title={Tensor field networks: Rotation-and translation-equivariant neural networks for 3d point clouds},
  author={Thomas, Nathaniel and Smidt, Tess and Kearnes, Steven and Yang, Lusann and Li, Li and Kohlhoff, Karthik and Riley, Patrick},
  journal={arXiv preprint arXiv:1802.08219},
  year={2018}
}

@inproceedings{fuchs2020se,
  title={{SE}(3)-transformers: 3d roto-translation equivariant attention networks},
  author={Fuchs, Fabian B and Worrall, Daniel E and Fischer, Volker and Welling, Max},
  booktitle={Advances in Neural Information Processing Systems},
  volume={33},
  pages={1970--1981},
  year={2020}
}

@inproceedings{maron2019universality,
  title={Universality of invariant networks},
  author={Maron, Haggai and Ben-Hamu, Heli and Shamir, Nadav and Lipman, Yaron},
  booktitle={International Conference on Machine Learning},
  pages={4363--4371},
  year={2019}
}

@inproceedings{anderson2019cormorant,
  title={Cormorant: Covariant molecular neural networks},
  author={Anderson, Brandon and Hy, Truong Son and Kondor, Risi},
  booktitle={Advances in Neural Information Processing Systems},
  volume={32},
  pages={14510--14519},
  year={2019}
}

@InProceedings{maron19a,
  title = 	 {On the Universality of Invariant Networks},
  author =       {Maron, Haggai and Fetaya, Ethan and Segol, Nimrod and Lipman, Yaron},
  booktitle = 	 {Proceedings of the 36th International Conference on Machine Learning},
  pages = 	 {4363--4371},
  year = 	 {2019},
  editor = 	 {Chaudhuri, Kamalika and Salakhutdinov, Ruslan},
  volume = 	 {97},
  series = 	 {Proceedings of Machine Learning Research},
  month = 	 {09--15 Jun},
  publisher =    {PMLR},
  url = 	 {https://proceedings.mlr.press/v97/maron19a.html}
}

@InProceedings{ravanbakhsh20a,
  title = 	 {Universal Equivariant Multilayer Perceptrons},
  author =       {Ravanbakhsh, Siamak},
  booktitle = 	 {Proceedings of the 37th International Conference on Machine Learning},
  pages = 	 {7996--8006},
  year = 	 {2020},
  editor = 	 {III, Hal Daumé and Singh, Aarti},
  volume = 	 {119},
  series = 	 {Proceedings of Machine Learning Research},
  month = 	 {13--18 Jul},
  publisher =    {PMLR},
  url = 	 {https://proceedings.mlr.press/v119/ravanbakhsh20a.html}
}

@inproceedings{
  song2021scorebased,
  title={Score-Based Generative Modeling through Stochastic Differential Equations},
  author={Yang Song and Jascha Sohl-Dickstein and Diederik P Kingma and Abhishek Kumar and Stefano Ermon and Ben Poole},
  booktitle={International Conference on Learning Representations},
  year={2021},
  url={https://openreview.net/forum?id=PxTIG12RRHS}
}

@inproceedings{
xie2024the,
title={The Price of Freedom: Exploring Tradeoffs between Expressivity and Computational Efficiency in Equivariant Tensor Products},
author={YuQing Xie and Ameya Daigavane and Mit Kotak and Tess Smidt},
booktitle={ICML 2024 Workshop on Geometry-grounded Representation Learning and Generative Modeling},
year={2024},
url={https://openreview.net/forum?id=0HHidbjwcf}
}

@inproceedings{bose2024sestochastic,
	author = {Joey Bose and Tara Akhound-Sadegh and Guillaume Huguet and Kilian Fatras and Jarrid Rector-Brooks and Cheng-Hao Liu and Andrei Cristian Nica and Maksym Korablyov and Michael M. Bronstein and Alexander Tong},
	booktitle = {The Twelfth International Conference on Learning Representations},
	title = {{SE}(3)-Stochastic Flow Matching for Protein Backbone Generation},
	url = {https://openreview.net/forum?id=kJFIH23hXb},
	year = {2024}}

@inproceedings{shi2021learning,
  title={Learning gradient fields for molecular conformation generation},
  author={Shi, Chence and Luo, Shitong and Xu, Minkai and Tang, Jian},
  booktitle={International conference on machine learning},
  pages={9558--9568},
  year={2021},
  organization={PMLR}
}

@inproceedings{kohler2020equivariant,
  title={Equivariant flows: exact likelihood generative learning for symmetric densities},
  author={K{\"o}hler, Jonas and Klein, Leon and No{\'e}, Frank},
  booktitle={International conference on machine learning},
  pages={5361--5370},
  year={2020},
  organization={PMLR}
}

@misc{levy2023using,
	archiveprefix = {arXiv},
	author = {Daniel Levy and S{\'e}kou-Oumar Kaba and Carmelo Gonzales and Santiago Miret and Siamak Ravanbakhsh},
	eprint = {2309.03139},
	primaryclass = {cs.LG},
	title = {Using Multiple Vector Channels Improves E(n)-Equivariant Graph Neural Networks},
	url = {https://arxiv.org/abs/2309.03139},
	year = {2023}}

@inproceedings{kim2023,
	author = {Kim, Jinwoo and Nguyen, Dat and Suleymanzade, Ayhan and An, Hyeokjun and Hong, Seunghoon},
	booktitle = {Advances in Neural Information Processing Systems},
	editor = {A. Oh and T. Naumann and A. Globerson and K. Saenko and M. Hardt and S. Levine},
	pages = {18582--18612},
	publisher = {Curran Associates, Inc.},
	title = {Learning Probabilistic Symmetrization for Architecture Agnostic Equivariance},
	url = {https://proceedings.neurips.cc/paper_files/paper/2023/file/3b5c7c9c5c7bd77eb73d0baec7a07165-Paper-Conference.pdf},
	volume = {36},
	year = {2023}}

@article{mondal2024equivariant,
	author = {Mondal, Arnab Kumar and Panigrahi, Siba Smarak and Kaba, Oumar and Mudumba, Sai Rajeswar and Ravanbakhsh, Siamak},
	journal = {Advances in Neural Information Processing Systems},
	title = {Equivariant adaptation of large pretrained models},
	volume = {36},
	year = {2024}}

@inproceedings{kaba2023equivariance,
	author = {Kaba, S{\'e}kou-Oumar and Mondal, Arnab Kumar and Zhang, Yan and Bengio, Yoshua and Ravanbakhsh, Siamak},
	booktitle = {International Conference on Machine Learning},
	organization = {PMLR},
	pages = {15546--15566},
	title = {Equivariance with learned canonicalization functions},
	year = {2023}}

@article{keriven2019universal,
	author = {Keriven, Nicolas and Peyr{\'e}, Gabriel},
	journal = {Advances in Neural Information Processing Systems},
	title = {Universal invariant and equivariant graph neural networks},
	volume = {32},
	year = {2019}}

@inproceedings{satorras2021n,
	author = {Satorras, V\'{\i}ctor Garcia and Hoogeboom, Emiel and Welling, Max},
	booktitle = {Proceedings of the 38th International Conference on Machine Learning},
	editor = {Meila, Marina and Zhang, Tong},
	month = {18--24 Jul},
	pages = {9323--9332},
	publisher = {PMLR},
	series = {Proceedings of Machine Learning Research},
	title = {E(n) Equivariant Graph Neural Networks},
	url = {https://proceedings.mlr.press/v139/satorras21a.html},
	volume = {139},
	year = {2021}}

@article{bloem2020probabilistic,
	author = {Bloem-Reddy, Benjamin and Teh, Yee Whye},
	journal = {The Journal of Machine Learning Research},
	number = {1},
	pages = {3535--3595},
	publisher = {JMLRORG},
	title = {Probabilistic symmetries and invariant neural networks},
	volume = {21},
	year = {2020}}

@misc{xu2022geodiffgeometricdiffusionmodel,
      title={GeoDiff: a Geometric Diffusion Model for Molecular Conformation Generation}, 
      author={Minkai Xu and Lantao Yu and Yang Song and Chence Shi and Stefano Ermon and Jian Tang},
      year={2022},
      eprint={2203.02923},
      archivePrefix={arXiv},
      primaryClass={cs.LG},
      url={https://arxiv.org/abs/2203.02923}, 
}

@misc{satorras2022enequivariantnormalizingflows,
      title={E(n) Equivariant Normalizing Flows}, 
      author={Victor Garcia Satorras and Emiel Hoogeboom and Fabian B. Fuchs and Ingmar Posner and Max Welling},
      year={2022},
      eprint={2105.09016},
      archivePrefix={arXiv},
      primaryClass={cs.LG},
      url={https://arxiv.org/abs/2105.09016}, 
}

@inproceedings{deng2021vector,
	author = {Deng, Congyue and Litany, Or and Duan, Yueqi and Poulenard, Adrien and Tagliasacchi, Andrea and Guibas, Leonidas J},
	booktitle = {Proceedings of the IEEE/CVF International Conference on Computer Vision},
	pages = {12200--12209},
	title = {Vector neurons: A general framework for so (3)-equivariant networks},
	year = {2021}}

@inproceedings{hoogeboom2022equivariant,
	author = {Hoogeboom, Emiel and Satorras, Victor Garcia and Vignac, Cl{\'e}ment and Welling, Max},
	booktitle = {International Conference on Machine Learning},
	organization = {PMLR},
	pages = {8867--8887},
	title = {Equivariant diffusion for molecule generation in 3d},
	year = {2022}}

@inproceedings{ravanbakhsh2017equivariance,
	author = {Ravanbakhsh, Siamak and Schneider, Jeff and Poczos, Barnabas},
	booktitle = {International Conference on Machine Learning},
	organization = {PMLR},
	pages = {2892--2901},
	title = {Equivariance through parameter-sharing},
	year = {2017}}

@article{qm9,
	author = {Raghunathan Ramakrishnan and Pavlo O. Dral and Matthias Rupp and O. Anatole von Lilienfeld},
	doi = {10.1038/sdata.2014.22},
	journal = {Scientific Data},
	month = {aug},
	number = {1},
	publisher = {Springer Science and Business Media {LLC}},
	title = {Quantum chemistry structures and properties of 134 kilo molecules},
	url = {https://doi.org/10.1038%2Fsdata.2014.22},
	volume = {1},
	year = 2014}

@inproceedings{schutt2017schnet,
	author = {Sch{\"u}tt, Kristof and Kindermans, Pieter-Jan and Felix, Huziel Enoc Sauceda and Chmiela, Stefan and Tkatchenko, Alexandre and M{\"u}ller, Klaus-Robert},
	booktitle = {Advances in Neural Information Processing Systems},
	pages = {991--1001},
	title = {Schnet: A continuous-filter convolutional neural network for modeling quantum interactions},
	year = {2017}}

@inproceedings{rotmnsit,
author = {Larochelle, Hugo and Erhan, Dumitru and Courville, Aaron and Bergstra, James and Bengio, Yoshua},
title = {An empirical evaluation of deep architectures on problems with many factors of variation},
year = {2007},
isbn = {9781595937933},
publisher = {Association for Computing Machinery},
address = {New York, NY, USA},
url = {https://doi.org/10.1145/1273496.1273556},
doi = {10.1145/1273496.1273556},
booktitle = {Proceedings of the 24th International Conference on Machine Learning},
pages = {473–480},
numpages = {8},
location = {Corvalis, Oregon, USA},
series = {ICML '07}
}

@inproceedings{e2cnn,
     title={{General {E(2)}-Equivariant Steerable CNNs}},
     author={Weiler, Maurice and Cesa, Gabriele},
     booktitle={Conference on Neural Information Processing Systems (NeurIPS)},
     year={2019},
}

@misc{ronneberger2015unetconvolutionalnetworksbiomedical,
      title={U-Net: Convolutional Networks for Biomedical Image Segmentation}, 
      author={Olaf Ronneberger and Philipp Fischer and Thomas Brox},
      year={2015},
      eprint={1505.04597},
      archivePrefix={arXiv},
      primaryClass={cs.CV},
      url={https://arxiv.org/abs/1505.04597}, 
}

@inproceedings{cohen2016group,
  title={Group equivariant convolutional networks},
  author={Cohen, Taco and Welling, Max},
  booktitle={Proceedings of the 33rd International Conference on Machine Learning (ICML)},
  pages={2990--2999},
  year={2016}
}

@inproceedings{cohen2018spherical,
  title={Spherical CNNs},
  author={Cohen, Taco S. and Geiger, Mario and K{\"o}hler, Jonas and Welling, Max},
  booktitle={International Conference on Learning Representations (ICLR)},
  year={2018}
}

@inproceedings{esteves2018learning,
  title={Learning SO(3) equivariant representations with spherical CNNs},
  author={Esteves, Carlos and Allen-Blanchette, Christine and Makadia, Ameesh and Daniilidis, Kostas},
  booktitle={Advances in Neural Information Processing Systems (NeurIPS)},
  pages={6541--6550},
  year={2018}
}

@inproceedings{finzi2020generalizing,
  title={Generalizing convolutional neural networks for equivariance to Lie groups on arbitrary continuous data},
  author={Finzi, Marc and Stanton, Samuel and Izmailov, Pavel and Wilson, Andrew Gordon},
  booktitle={Proceedings of the 37th International Conference on Machine Learning (ICML)},
  pages={3165--3176},
  year={2020}
}

@inproceedings{weiler20183d,
  title={3D steerable CNNs: Learning rotationally equivariant features in volumetric data},
  author={Weiler, Maurice and Cesa, Gabriele},
  booktitle={Advances in Neural Information Processing Systems (NeurIPS)},
  pages={10402--10413},
  year={2018}
}

@inproceedings{weiler2019general,
  title={General {E(2)}-equivariant steerable {CNNs}},
  author={Weiler, Maurice and Cesa, Gabriele},
  booktitle={Advances in Neural Information Processing Systems (NeurIPS)},
  pages={14334--14345},
  year={2019}
}

@inproceedings{worrall2017harmonic,
  title={Harmonic networks: Deep translation and rotation equivariance},
  author={Worrall, Daniel E. and Garbin, Stephan J. and Turmukhambetov, Daniyar and Brostow, Gabriel J.},
  booktitle={Proceedings of the IEEE Conference on Computer Vision and Pattern Recognition (CVPR)},
  pages={5028--5037},
  year={2017}
}

@article{brandstetter2022geometric,
  title={Geometric deep learning on molecular representations},
  author={Brandstetter, Johann and Grisafi, Andrea and Ceriotti, Michele and Corminboeuf, Cl{\'e}mence and Rupp, Matthias},
  journal={arXiv preprint arXiv:2206.03083},
  year={2022}
}

@article{schutt2021equivariant,
  title={Equivariant message passing for the prediction of tensorial properties and molecular spectra},
  author={Sch{\"u}tt, Kristof T. and Gastegger, Michael and Tkatchenko, Alexandre and M{\"u}ller, Klaus-Robert and Maurer, Reinhard J.},
  journal={arXiv preprint arXiv:2102.03150},
  year={2021}
}

@article{gilmer2017neural,
  title={Neural message passing for quantum chemistry},
  author={Gilmer, Justin and Schoenholz, Samuel S. and Riley, Patrick F. and Vinyals, Oriol and Dahl, George E.},
  journal={arXiv preprint arXiv:1704.01212},
  year={2017}
}

@article{unke2019physnet,
  title={PhysNet: A neural network for predicting energies, forces, dipole moments, and partial charges},
  author={Unke, Oliver Thorsten and Meuwly, Markus},
  journal={Journal of Chemical Theory and Computation},
  volume={15},
  number={6},
  pages={3678--3693},
  year={2019}
}

@article{zhou2020group,
  title={Group-Theory Based Decomposition of Convolutional Networks},
  author={Zhou, Yanzhi and Feng, Jiashi},
  journal={Proceedings of the IEEE Conference on Computer Vision and Pattern Recognition (CVPR)},
  year={2020},
  pages={6530--6539}
}

@article{falorsi2018explorations,
  title={Explorations in Homeomorphic Variational Auto-Encoding},
  author={Falorsi, Luca and de Haan, Pim and Davidson, Tim R. and De Cao, Nicola and Weiler, Maurice and Forr{\'e}, Patrick and Cohen, Taco S.},
  journal={arXiv preprint arXiv:1807.04689},
  year={2018}
}

\appendix

\section{Related Work}
\label{apd:background}

\paragraph{Equivariant neural networks}

Equivariant neural networks have been successfully applied in fields such as image processing~\cite{cohen2016group, worrall2017harmonic, weiler2019general}, 3D data processing~\cite{thomas2018tensor, fuchs2020se, weiler20183d, deng2021vector, esteves2018learning}, graph analysis~\cite{maron2019universality, keriven2019universal}, physical simulations~\cite{finzi2020generalizing, brandstetter2022geometric}, and AI for science applications such as molecular modeling and computational chemistry~\cite{schutt2017schnet, anderson2019cormorant, schutt2021equivariant, satorras2021n, gilmer2017neural, unke2019physnet}, leading to improved generalization and sample efficiency. Despite their benefits, existing techniques to build equivariant neural networks face several challenges. One significant issue is \emph{\textbf{computational complexity}}, as specialized operations like group convolutions or spherical harmonics are computationally expensive~\cite{fuchs2020se, kondor2018clebsch, weiler2019general, cohen2018spherical}. Additionally, there are \emph{\textbf{expressivity constraints}}, since limiting models to equivariant functions may restrict their ability to capture complex patterns not perfectly aligned with the symmetry group~\cite{maron2019universality, ravanbakhsh2017equivariance, zhou2020group}. Moreover, the design of these networks inherently presents \emph{\textbf{significant complexities and challenges}}, necessitating tailored architectural solutions that add layers of difficulty to both development and practical implementation.

\paragraph{Canonicalization for architecture-agnostic equivariance}

Canonicalization offers an alternative approach to designing specialized neural networks by transforming data into a standardized or ``canonical" form, effectively removing symmetry-related variability. The canonicalization function maps each data point to a representative element in its orbit under the group. Recent works propose learning the canonicalization function jointly with the main task-specific network \cite{kaba2023equivariance, kim2023, mondal2024equivariant, allingham2024generative}. By parameterizing $c$ as a neural network, the model can discover an optimal way to standardize data, leading to improved representations and performance.

\paragraph{Equivariance and Diffusion Models}

Diffusion models have emerged as a powerful class of generative models, achieving remarkable success in generating high-fidelity samples across various domains~\cite{sohl2015deep, ho2020denoising, song2021scorebased}. These models learn data distributions by reversing a diffusion process that incrementally adds Gaussian noise to the data. 
Incorporating equivariance into diffusion models can enhance their performance on data with inherent symmetries~\cite{hoogeboom2022equivariant, xu2022geodiff, falorsi2018explorations}. Equivariant diffusion models ensure that the generative process respects the symmetries of the data, leading to more efficient learning and better generalization. However, integrating equivariance into diffusion models also inherits the challenges associated with equivariant neural networks.

\section{Experimental Details}\label{apd:first}

\subsection{RotMNIST Dataset}
A denoising U-Net is trained with a base dimension of 64. The discrete image canonicalizer is implemented using EquiAdapt \citep{equiadapt}. An Equivariant Steerable Neural Network (ESCNN) \citep{e2cnn} with 16 channels, 5 layers and kernel size 5 is used for canonicalization. Both networks are trained for 1000 epochs on an NVIDIA A100-SXM4-80GB over roughly 9 hours.
\begin{table}[!htbp]
\centering
\caption{RotMNIST Results: reconstruction L2 distance of denoising model over the test set with randomly sampled noising timesteps.}
\label{tab:rotmnist}
\begin{tabular}{lr}
\toprule
                         Model &  Test Reconstruction Loss \\
\midrule
UNet &        0.0312 \\
canon(p32)+UNet &        0.0264 \\
canon(p16)+UNet &        \textbf{0.0255} \\
canon(p8)+UNet &        0.0264 \\
canon(p16, fr)+UNet & 0.0264 \\
\bottomrule
\end{tabular}
\end{table}

\subsection{QM9 Dataset}
The Equivaraint Diffusion Model (EDM) and Graph Diffusion Model (GDM) are trained with hyperparameters equivalent to \citet{hoogeboom2022equivariant}, namely 1000 diffusion steps with polynomial noise schedule and precision 1e-5. An L2 denoising loss is used with batch size 64. The denoiser has 256 node features and 9 layers. An EMA decay of 0.9999 is used. For the canonicalization network, a multi-channel EGNN \citep{levy2023using, satorras2021n} is used to output two rotation vectors which are then orthonomalized into a rotation matrix via modified Gram-Schmidt. The network has 9 layers and 64 node features. A single run on a Quadro RTX 8000 takes roughly 4 days.

\end{document}